\documentclass[10pt,twocolumn,letterpaper]{article}

\usepackage[pagenumbers]{wacv} 

\usepackage{graphicx}
\usepackage{amsmath}
\usepackage{amssymb}
\usepackage{booktabs}
\usepackage{pifont}
\newcommand{\cmark}{\ding{51}}%
\newcommand{\xmark}{\ding{55}}%
\usepackage[title]{appendix}

\usepackage[pagebackref,breaklinks,colorlinks]{hyperref}

\usepackage[capitalize]{cleveref}
\crefname{section}{Sec.}{Secs.}
\Crefname{section}{Section}{Sections}
\Crefname{table}{Table}{Tables}
\crefname{table}{Tab.}{Tabs.}

\begin{document}

\title{Interpretable Convolutional SyncNet}

\author{Sungjoon Park\\
Hudson-AI \\
{\tt\small sungjoon.park@hudson-ai.com}
\and
Jaesub Yun\\
Hudson-AI \\
{\tt\small jaesub.yun@hudson-ai.com}
\and
Donggeon Lee\\
Hudson-AI \\
{\tt\small donggeon.lee@hudson-ai.com}
\and
Minsik Park\\
Hudson-AI \\
{\tt\small mspark@hudson-ai.com}
}
\maketitle

\begin{abstract}
    Because videos in the wild can be out of sync for various reasons, a sync-net is used to bring the video back into sync for tasks that require synchronized videos.
    Previous state-of-the-art (SOTA) sync-nets use InfoNCE loss, rely on the transformer architecture, or both.
    Unfortunately, the former makes the model's output difficult to interpret, and the latter is unfriendly with large images, thus limiting the usefulness of sync-nets.
    In this work, we train a convolutional sync-net using the balanced BCE loss (BBCE), a loss inspired by the binary cross entropy (BCE) and the InfoNCE losses.
    In contrast to the InfoNCE loss, the BBCE loss does not require complicated sampling schemes.
    Our model can better handle larger images, and its output can be given a probabilistic interpretation.
    The probabilistic interpretation allows us to define metrics such as probability at offset and offscreen ratio to evaluate the sync quality of audio-visual (AV) speech datasets.
    Furthermore, our model achieves SOTA accuracy of $96.5\%$ on the LRS2 dataset and $93.8\%$ on the LRS3 dataset.
\end{abstract}

\label{sec:intro}
\section{Introduction}
The question of whether a video is in sync is, at first glance, an unexciting problem.
Delving deeper, one finds that it is closely related to many of the problems that are fundamental in machine learning: it is not only a classification task, but also a representation learning task on a multimodal dataset.
For an audio-visual (AV) speech dataset, the images are either in-sync or out-of-sync with the audio, which makes this a binary classification problem.
This problem is usually solved by encoding the images and audio into a common embedding space and comparing their similarity, which is a popular approach to representation learning on multimodal datasets.
A model trained on the AV synchronization task can be used to determine whether a video is in sync for some downstream task \cite{chung2017lip,son2017lip,afouras2018deep}, or can be used as a pre-trained model for various other tasks \cite{afouras2020self,cheng2020look,korbar2018cooperative,owens2018audio,patrick2020multi}.

Recent developments in sync-nets, by which we refer to various models that solve the problem of AV synchronization, have been dominated by the transformer architecture \cite{vaswani2017attention}.
This is natural since transformers are known to perform well on sequential data.
However, transformer based sync-nets are not suitable for some of the tasks for which sync-nets are useful for.
Speech-to-lip synthesis is one such task, where one generates image sequences of the face that is in sync with the input audio.
For this task, sync-nets are often used to help generate lip motions that are in sync with the input speech by penalizing generated images with unnatural lips through the sync-loss.
To our knowledge, all of the currently proposed transformer based sync-nets for AV speech datasets either take small images as inputs or perform aggressive downsampling.
Thus, much of the detailed spatial information that is useful for constraining the lip motion gets lost, rendering these transformer based sync-nets unsuitable for computing the sync-loss.

Another problem is that sync-nets are often trained using the InfoNCE loss \cite{oord2018representation} to select the in-sync AV pair out of an in-sync AV pair and multiple out-of-sync AV pairs.
This task aligns well with the conventional evaluation scheme of sync-nets, where one records the average probability of correctly identifying an in-sync AV pair out of an in-sync and multiple out-of-sync AV pairs.
Unfortunately, such a model's output is hard to interpret because it is a relative evaluation of AV sync, and although it is possible to use the InfoNCE loss as a sync-loss by brute-force implementation~\cite{zhou2021pose,wang2023progressive}, it requires sampling many out-of-sync pairs, and the sampling scheme used (such as how the negative samples are chosen) when computing the sync-loss in speech-to-lip synthesis should also match the sampling scheme used during the training of the sync-net.
For these reasons, works on speech-to-lip synthesis that utilize sync-nets often train their own convolutional neural net (CNN) based sync-nets using the binary cross entropy (BCE) loss \cite{prajwal2020lip,park2022synctalkface,yin2022styleheat,wang2022one}.
The convolutional nature of these models better preserves the spatial information, and the BCE loss used to train the sync-net can also be used as a sync-loss without complications.

In this work, we train a powerful interpretable and convolutional sync-net (IC-SyncNet) that achieves SOTA accuracy of $96.5\%$ and $93.8\%$ on the LRS2 \cite{afouras2018deep} and LRS3 \cite{afouras2018lrs3} test datasets, respectively.
Our model takes five $128 \times 256$ images and $32 \times 80$ mel spectrogram as inputs, and has 40.6M parameters.
Since there is no aggressive pooling, spatial information is better preserved than the transformer-based models.
Furthermore, our loss function, which we call the balanced BCE (BBCE) loss, is designed to mimick the InfoNCE loss but is built using the BCE loss.
We find that unlike the InfoNCE loss, the BBCE loss does not require complicated sampling schemes to train.
The difference is manifest for the LRS3 dataset, and we attribute this to the fact that these two losses maximize lower bounds on different mutual informations.
Furthermore, the BCE nature of the BBCE loss allows us to interpret the model's output as the probability that the video is in sync, which simplifies some of the complications that follow the InfoNCE.
To demonstrate the usefulness of the probabilistic interpretation, we show how the IC-SyncNet can be used to detect active speakers.
Using the insights from this application, we propose to use offset, probability at offset, and offscreen ratio as metrics to evaluate the sync-qualities of various AV speech datasets.

A summary of the main contributions is as follows:
\begin{itemize}
    \item \textbf{We propose to use the BBCE loss.} BBCE and InfoNCE show similar performance but BBCE output is much easier to interpret. 
    \item \textbf{We formulate sync quality metrics of talking heads, which we call offset, probability at offset, and off-screen ratio.} These metrics can be used to filter out bad data that is common in talking head datasets.
    \item \textbf{We obtained SOTA results with simple CNN.}
\end{itemize}

\section{Related Works}
\label{sec:related_works}
\subsection{Audio-visual speech synchronization}
The objective of AV synchronization to determine whether image and audio sequence are in sync.
We will refer to any model that solves this problem as a sync-net.
We always assume that videos are set to 25 fps.

\textbf{SyncNet.}
The first convolutional sync-net was proposed in \cite{chung2017out}.
The authors train two encoders, each for the visual and audio stream.
The visual encoder takes 5 consecutive grayscaled images of the mouth, which is encoded into $v \in \mathbb{R}^{256}$.
Similarly, the audio encoder takes the mel frequency ceptral coefficients (MFCC) of the corresponding audio segment, which is also encoded into $a \in \mathbb{R}^{256}$.
The two encoders are trained using the contrastive loss \cite{chopra2005learning}
\begin{equation}
\mathcal{L}_{\textrm{SyncNet}} = \frac{1}{2B} \sum_{n=1}^{B} \left[ y_n d_n^2 + (1-y_n) \max(m-d_n,0)^2\right], \label{eq:loss_syncnet}
\end{equation}
where $d_n = \| v_n - a_n\|_2$, $m$ is the margin, $B$ is the batch size, and $y \in \{ 0,1\}$ indicates whether the video is in sync.
This model is usually referred to as the SyncNet.

\textbf{PM (Perfect Match).}
In \cite{chung2019perfect}, the authors train convolutional visual and audio encoders using 5 consecutive (color) images and the MFCC of the corresponding audio segment.
A major difference to the SyncNet is the usage of multi-way classification loss
\begin{equation}
\mathcal{L}_{\textrm{PM}} = -\frac{1}{B}\sum_{n=1}^{B} \ln \frac{\exp{1/d_{n,1}}}{\sum_{i=1}^{N} \exp{1/d_{n,i}}}, \label{eq:loss_pm}
\end{equation}
where $d_{n,i}=\|v_{n}-a_{n,i}\|_2$.
Here, $n$ is the index of sample inside the batch, and $i$ is the index of $N$ audio samples, where $i=1$ corresponds to the positive (i.e. in-sync with $v$) audio sample and $i=2,...,N$ correspond to negative (i.e. out-of-sync with $v$) audio samples.

\textbf{AVST.} 
In AVST \cite{chen2021audio}, the authors make use of the transformer architecture.
First, two CNNs are used to extract features from the image sequence and the audio (linear spectrogram), which are denoted as $v_i\in \mathbb{R}^{c \times t_v}$ and $a_j \in \mathbb{R}^{c \times t_a}$ respectively.
Here, $c$ is the channel dimension, $t_{v}$ and $t_{a}$ are the number of indices in the time dimension for the image and audio sequences, and $i$ and $j$ are the indices of samples inside the batch.
Then, $t_{v}$ and $t_{a}$ are fed into a cross-modal transformer model, by which we mean a transformer model that computes intra-modal and inter-modal attentions.
The output $s_{ij}$ corresponding to the [CLS] token is used as the similarity measure between audio and visual data, and the model is trained on the InfoNCE loss \cite{oord2018representation}
\begin{equation}
    \mathcal{L}_{\textrm{AVST}} = -\frac{1}{B} \sum_{i=1}^{B} \ln \frac{\exp{s_{ii}}}{\sum_j \exp{s_{ij}}}.\label{eq:loss_avst}
\end{equation}

\textbf{VocaLiST.}
VocaLiST \cite{kadandale2022vocalist} also uses a cross-modal transformer architecture after extracting image and audio features using CNNs.
A major difference to the AVST is that here, the authors max-pool the output of the cross-modal transformer in the temporal direction, which is then passed through a tanh layer and a fully connected layer to obtain the logits for the probability $p_n$ that the image sequence and the audio (mel spectrogram) are in sync. 
The BCE loss
\begin{equation}
    \mathcal{L}_{\textrm{VocaLiST}} = -\frac{1}{B}\sum_{n=1}^{B} \left[ y_n \ln p_n + (1-y_n) \ln(1-p_n)\right], \label{eq:loss_vocalist}
\end{equation}
is used to train the model to distinguish in-sync and out-of-sync image sequences and mel spectrograms, which are sampled equally.

\textbf{ModEFormer.} In \cite{gupta2023modeformer}, instead of using a cross-modal transformer, the authors utilize two unimodal transformers.
First, the image sequence is passed through a CNN, whose output is then fed into a (unimodal) transformer, and similarly for the corresponding mel spectrogram. The output corresponding to the [CLS] token for image sequence ($v$) and mel spectrogram ($a$) are used to train the model with the following variant of the InfoNCE loss,
\begin{align}
    \mathcal{L}_{\textrm{ModEFormer}} &= -\frac{1}{B} \sum_{v,a^+ \in \mathcal{P}} \ln \frac{\exp(\phi(v,a^{+})/\tau)}{\sum_{a \in \mathcal{N}(v)} \exp(\phi(v,a)/\tau)} \label{eq:loss_modeformer}\\
    \phi(v,a) &= \frac{v}{\|v\|_2} \cdot \frac{a}{\|a\|_2}. \label{eq:cossim}
\end{align}
Here, $\tau$ is the temperature parameter, $\mathcal{P}$ is a set of positive pairs (i.e. in-sync), and $\mathcal{N}(v)$ is a set of random negative audio samples along with the positive audio sample. 
These negative audio samples can be from the same video (hard negatives) or a different video (easy negatives).
A two-stage training procedure is used, where the model is trained with just the hard negatives during the first stage, and then with both hard and easy negatives during the second stage.
This method produces the currently best performing model.

\subsection{Speech-to-lip synthesis}
The goal of speech-to-lip synthesis is to produce realistic talking face images that match the speech signal.
Although various methods have been proposed, a form of sync-loss \cite{prajwal2020lip,park2022synctalkface,zhang2023dinet,muaz2023sidgan,guan2023stylesync,zhou2021pose,wang2023progressive}, is often used to enforce the generator network to synthesize lip shapes that are in sync with the speech signal. 
To compute the sync-loss, previous transformer based sync-nets are unsuitable for two reasons. 
First, the images that are fed into the transformer based models are either low in resolution (e.g. VocaLiST and ModEFormer use input size of 48 x 96), or downsampled too aggressively (e.g. AVST uses input size of 224 x 224, but carries out max pooling at 14 x 14 resolution, which destroys spatial information).
Second, these transformer models are heavy, which can limit the batch size when training the speech-to-lip generator.
There does not seem to be a simple solution for these architectures, which use CNN followed by transformer.

It is simpler to train a reasonably light-weight but powerful convolutional sync-net.
This is an important problem in the field because a sync-net that is of poor quality will only be able to give poor guidance to the generator.
However, most of the works, with the exception of \cite{muaz2023sidgan} which explores the effect of restoring translational invariance in sync-nets, do not delve deeply into the accuracy of the sync-nets.
In this work, we propose a recipe that can be used to train powerful convolutional sync-nets that is also suitable for speech-to-lip generation.

\subsection{Representation learning}
Broadly speaking, AV speech synchronization is also related to the field of representation learning, such as self-supervised learning.
The goal of self-supervised learning is to train models to learn meaningful features without supervision on unlabeled data. 
These pre-trained models can then be used as a starting point for various other tasks.
Models pre-trained with methods such as SimCLR \cite{chen2020simple} perform as well as models pre-trained with supervision, and more recent methods can outperform supervised pre-training, see \cite{balestriero2023cookbook} and references therein.

Because the problem of AV speech synchronization is a multimodal task, representation learning methods that use multimodal data are more closely related to the problem we consider.
For example, for multimodal data consisting of images with captions, one has access to two different views of the same data, which come in the form of image and text.
This provides a natural means to train models with contrastive learning as in CLIP \cite{radford2021learning}.
Similarly, for video data, one has access to images and the corresponding audio, so that with a suitable pretext task, one can train powerful encoders such as AV-HuBERT \cite{shi2022learning}.

\section{Our Proposed Training Method}
Our method is designed such that (1) spatial information is well-preserved (Sec. \ref{ssec:architecture},\ref{ssec:drop_and_tune}) (2) model's output is interpretable (Sec. \ref{ssec:bbce_loss}) (3) there are no unnecessary information bottlenecks (Sec. \ref{ssec:data_processing}).

\subsection{Architecture}
\label{ssec:architecture}
\begin{figure}
	\centering
	\includegraphics[width=8cm]{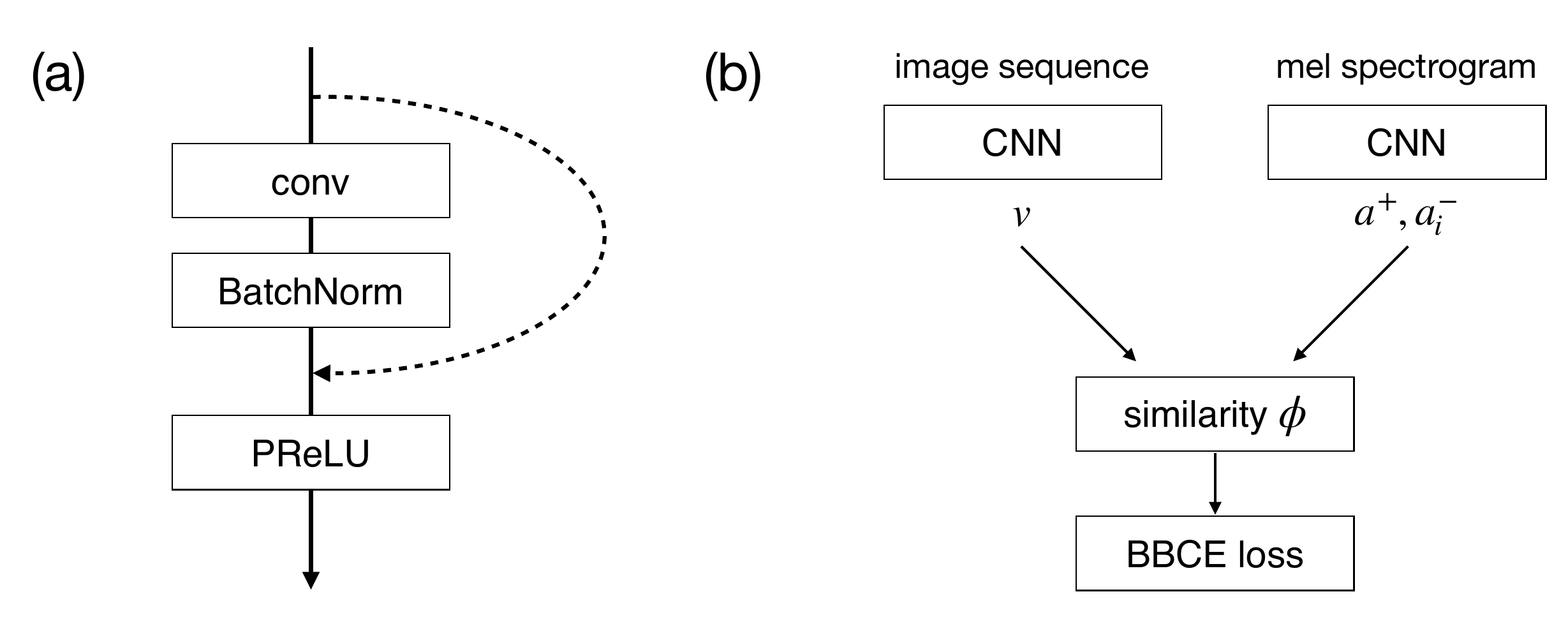}
	\caption{(a) Convolution block used in the encoders. (b) Summary of training our sync-net.}
	\label{fig:fig1}
\end{figure}

\textbf{Backbone.}
We use simple CNN encoders for both image sequence and mel spectrogram.
Our convolutional block is composed of Conv-BN-PReLU with skip connection, see Figure~\ref{fig:fig1} (a). 
We simply stack these blocks to obtain encoders for both image sequences and mel spectrograms.
We define the mouth region as the lower-half of the detected face.
Since we use five consecutive images stacked along the channel dimension, as is conventional, the input to the visual encoder is $15 \times H \times W$, where $H=128$ and $W=256$.
The dimension of the mel spectrogram is $32 \times 80$, see Sec. \ref{ssec:data_processing}.
Both the images and the mel spectrograms are embedded into $\mathbb{R}^{512}$.

\textbf{Anti-aliasing.}
As mentioned above, we use five consecutive images of the mouth region as the input to the visual encoder.
Although the inputs have mouth roughly aligned at the center, face detectors used to produce the bounding box are not perfect.
In particular, there can be translational and scale noise to the face bounding box.
Since pooling (stride in convolution) destroys translational invariance, we use BlurPool when reducing the spatial dimensions for the visual encoder~\cite{zhang2019making}.

\textbf{DropBlock.}
We would like our model to fully utilize the information contained in the images and audio to determine whether they are in sync.
To make sure that the model learns to look evenly at the input information, we insert DropBlock layers into our CNN, which is known to help the model to learn more spatially distributed representations \cite{ghiasi2018dropblock}.
As in the original implementation, we use two DropBlock layers, each with $10\%$ drop rate.
For the audio encoder, we use the 2D DropBlock, but for the visual encoder, we use the 3D DropBlock so that blocks are dropped in the channel dimension as well as the spatial dimensions.
This is to make sure that the model learns to look more evenly at the temporal information.

\textbf{Drop-and-tune strategy.}
\label{ssec:drop_and_tune}
It is well-known that the simultaneous usage of Dropout \cite{srivastava2014dropout} and batch normalization (BN) \cite{ioffe2015batch} can result in lower performance because Dropout causes a shift in the input statistics of the BN layer between the train and the test phase \cite{li2019understanding}.
Although it is usually not mentioned in the literature, DropBlock can cause a shift in the input statistics of the BN layer for the same reasons.
Indeed, we did not observe any performance boost by simply inserting DropBlock layers into our model.

Since it is the change in the input statistics that is causing the problem, we can cast this as a domain adaptation problem.
One simple solution to the problem of domain shift is to adopt the BN statistics of the data in the new domain \cite{li2016revisiting}, a method that was also mentioned in \cite{li2019understanding}.
We take a step further and add an additional fine-tuning phase, where the drop rate is set to zero and only the parameters in the BN layers are trained.
Since the convolutional weights are already regularized to encode more evenly distributed information, this gives a simple way to keep the benefits of both the BN and DropBlock.
We found that this gives a significant boost to the accuracy.

\subsection{BBCE loss}
\label{ssec:bbce_loss}
\textbf{Parameterizing the probability.}
Denoting by $v$ and $a$ the embeddings of the images and mel spectrogram, we use the cosine similarity as a measure of the distance between them.
The probability that $v$ and $a$ are in sync is computed as
\begin{equation}
    p = \sigma (\phi(v,a)/\tau), \label{eq:sync_prob}
\end{equation}
where $\sigma$ is the sigmoid, $\phi$ is the cosine similarity \eqref{eq:cossim}, and $\tau$ is the temperature. 
Note that when using the cosine similarity, we can assume that $v$ and $a$ lie on the unit sphere.
This is a common practice and has the advantage that the space in which the embeddings live is bounded, which helps with stabilty in training.
However, it is not a good idea to use $p = \sigma (\phi(v,a))$ for two reasons. First, $-1 \le \phi(v,a) \le 1$ so that $0.269 \approx \sigma(-1) \le p \le \sigma(1) \approx 0.731$ (not suitable as a parameterization of probability). 
Second, if we use the BCE loss without $\tau$, even after the model has learned to pull positive pairs together and negative pairs apart in the optimal way by considering all of the data, the loss will not saturate for a batch of samples.
This is because for a positive (negative) pair, the loss is minimized when the cosine similarity is $1$ ($-1$).
However, it is impossible to embed all of the data into this configuration: for a single $v$, all out-of-sync $a$ must lie on the oppsite pole in the hypersphere, which is surely not optimal for a different $v$.
Thus, even if we have found the globally optimal embeddings for the dataset, the gradient of a minibatch at this configuration can be significant.
We note that many sync-nets suffer from analogous problems arising from parameterizations that are not well-motivated \cite{zhou2021pose,wang2023progressive,prajwal2020lip,park2022synctalkface,wang2022one}.

On the other hand, if we have a small temperature dividing the cosine similarity, it is not necessary for the angle between audio and visual embeddings to be large for the loss to saturate (i.e. give small gradients).
Thus, we use a learnable temperature parameter in the spirit of CLIP \cite{radford2021learning} to let the system figure out the optimal scaling of the similarity.
As expected, the temperature and the magnitudes of consine similarities gradually decreases during the training.
At the end of the training on the LRS2 dataset, the cosine similarities of positive and negative pairs reach an average of approximately $0.1$ and $-0.18$.
This indicates that the model can classify negative pairs with a larger margin.
One might therefore expect that the negative pairs are easier to learn, and therefore, the number of false positives should be smaller than the number of false negatives.
Interestingly, we find the opposite to be true: the number of false positives is usually greater than the number of false negatives.
This seemingly paradoxical phenomenon is related to the presence of the silent regions during which nobody is speaking, see the Appendix for further details.

\textbf{Loss function.}
Although the InfoNCE loss is a popular choice, it is not without disadvantages.
For example, directly implementing InfoNCE as the sync-loss during the training of speech-to-lip synthesis can consume significant resources (can require many negatives) and the loss can be sensitive to the distribution of the negatives.
Furthermore, using the BCE as the sync-loss (\cite{jang2023s} does something similar) for a model trained on the InfoNCE loss is not well-motivated, unlike for a model trained on the BBCE loss which we will soon define.
We therefore turn to the classification aspect of the AV synchronization problem.
Our initial choice was to use the BCE loss as in \cite{kadandale2022vocalist}, which does not suffer from the above problems.
However, a drawback of the ordinary BCE loss is that an image sequence is only compared with a single mel spectrogram, which is randomly chosen to be positive or negative.
Thus, a model that has seen a positive pair from a video must wait until we have processed through the whole dataset to see a different pair from the same video which could be either positive or negative.
A simple fix is to include a positive pair and a negative pair from the same video in a single batch, or better yet, we can include a single positive pair and multiple negative audio pairs, similarly to the InfoNCE loss.
We use multiple negative audio pairs because it is more memory efficient: mel spectrograms ($32 \times 80$) are much smaller than images ($128 \times 256 \times 3$).
Note that the negative audio pairs can come from the same video (hard negatives) or a different video (easy negatives).
Combining these possibilities, we obtain the balanced binary cross entropy (BBCE) loss:
\begin{align}
    \mathcal{L}_{\textrm{BBCE}} = -\frac{1}{2B} \sum_{n=1}^{B} &\Bigg[\ln p_{n,1} + \frac{1-w_e}{N_h}\sum_{i=1}^{N_h} \ln (1-p_{n,1+i}) \nonumber \\
    &+ \frac{w_e}{N_e} \sum_{i=1}^{N_e} \ln (1-p_{n,1+N_h+i})\Bigg].
\end{align}
Here, $p_{n,i}$ is the probability as parameterized in Eq.~\eqref{eq:sync_prob} that $v_n$ and $a_{n,i}$ is a positive pair (we assume that $v_n$ and $a_{n,1}$ is a positive pair, and the rest are negative pairs), $N_h$ is the number of hard negatives, $N_e$ is the number of easy negatives, and $w_e$ is the weight given to the easy negatives.
Note that the weights given to the negatives sum to one in order to simulate a balanced dataset between the positives and negatives.
The gradient from this loss is analogous to that from the InfoNCE loss as shown in the Appendix, but the model's output can be related to the probability of video sync through Eq.~\eqref{eq:sync_prob}.
Furthermore, this loss does not require two-stage training to get optimal performance as in ModEFormer, which uses InfoNCE.
This is discussed further in Sec.~\ref{ssec:cf_infonce}.

\subsection{Data processing}
\label{ssec:data_processing}
\textbf{Augmentation.}
Since the output of a face detector contains translational and scale noise, we use frame-wise random translation and scale augmentation. 
We also perform random horizontal flip on the image sequence.

\textbf{Audio window.}
Following previous studies, we sample images at 25 fps and audio at 16,000 Hz.
The audio is encoded into the mel spectrogram, which is computed using window size of 800, hop size of 200, and 80 frequency bins.
For the image sequence, we follow the convention of using 5 consecutive images.
Because $5/25=0.2$ and $16*200/16000 = 0.2$, the convention is to use $16 \times 80 $ mel spectrogram for the audio corresponding to the image sequence, where the first frame of the mel spectrogram is temporally aligned with the first frame of the image sequence.
However, this configuration causes information bottleneck in the sense that the mutual information between the image sequence and the audio sequence is bottlenecked by the limited window of the audio sequence.
A simple way to understand this is to notice that viseme formation is instantaneous in the sense that it is contained within a single image, whereas phoneme formation is not contained in a single time slice of the mel spectrogram.
Because the formation of phonemes and its relation to the visible information (such as visemes) is complex, and also depends on the language as well as the individual, it is difficult to give precisely what the optimal audio window should be.
Since the order of magnitude of the duration of a phoneme is around 0.1s for fast speech \cite{kuwabara1996acoustic}, giving an image a $0.2$s worth of audio sequence with the image at the center is a reasonable choice.
Therefore, we increase the audio window by $0.1$s on each side, so that a total of $0.4$s of audio is matched with 5 images.
This is the reason that we use 32 x 80 mel spectrograms as inputs to the audio encoder.
A practical side to this modification is that the size of a mel spectrogram is much smaller compared to an image, so that the increase in the accuracy comes at a cheap price.

\section{Sync Quality Metrics}
\label{sec:av_speech_dataset_quality}

\begin{figure*}
	\centering
	\includegraphics[width=12cm]{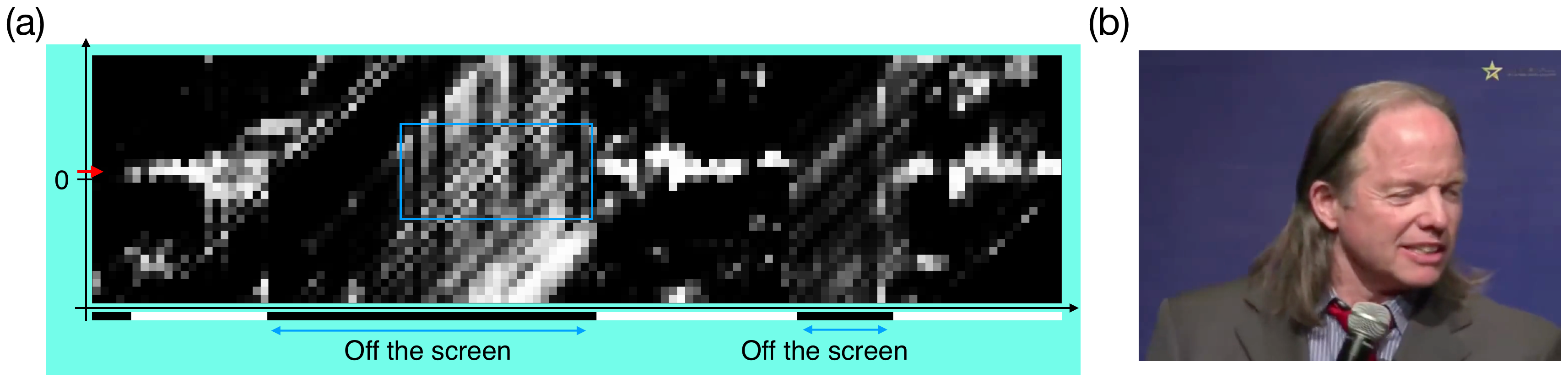}
	\caption{
\textbf{Active speaker detection} (from AVSpeech dataset \cite{cheng2020look}). 
(a) The probability of synchronization computed using a model trained on the LRS2 dataset. The x axis is the time step for images, and the y axis is the offset of audio with respect to the images. Offset for this video is indicated with red arrow.
(b) The person caught on screen.
The probability at offset is $0.29$ and the offscreen ratio is $0.48$.
}
	\label{fig:fig2}
\end{figure*}
To motivate the definition of quality metrics, let us first consider the problem of active speaker detection.
Recall that the cosine similarity of models trained using the BBCE loss has the interpretation as the model's confidence of whether the audio and images are in sync through Eq.~\eqref{eq:sync_prob}.
This makes the cosine similarity of the model trained using the BBCE loss more useful than that of the InfoNCE loss.
For example, it can easily be deployed for active speaker detection: for an in-sync video, the cosine similarity is positive when the audio matches the person's mouth movements, but it becomes negative when someone else is speaking.
This is because the positive and negative samples are balanced during training, so that for our purposes, the model output is well calibrated in the sense that $p=0.5$ is an appropriate tipping point for determining the sync.

As an example, we visualize the probability of video and audio being in sync in Fig.~\ref{fig:fig2} for various offsets of audio with respect to image sequence.
We find that the offset for this video is 1, meaning that audio starts about $1/25$ seconds before the images.
The probabilities at this offset is nearly 1 for the duration when the person on the screen is speaking. 
However, as soon as someone off the screen speaks, the probability becomes small, as expected.
We also find regions where the probability fluctuates, as indicated by a blue rectangle.
A possible explanation is that the person captured maintains the mouth shape as shown in (b), which can partially explain the audio for the duration seen by the sync-net.
Of course, on average, the probability of the audio and image sequences being in sync is not large, and can be detected through a combination of smoothing and thresholding, see the Appendix for details.
A prediction of active speaker through such an algorithm is shown below the probability map in Fig.~\ref{fig:fig2}.

After examining multiple samples from various AV speech (talking head) datasets, we have identified three frequent classes of problems: (1) the video is not in sync (2) someone off-the-screen is speaking as discussed above (3) the overall sync quality of video is poor (e.g. stuttering videos from video conferencing).
Therefore, we use the offset, offscreen ratio, and overall sync confidence to evaluate the quality of AV speech data to quantify the above problems.
The offscreen ratio is computed by finding portions of the video where the sync quality is poor as discussed above, and the overall sync quality is found by computing the probability of the average cosine similarity at the offset.
A detailed evaluation of the sync qualities for LRS2 \cite{afouras2018deep}, LRS3 \cite{afouras2018lrs3}, AVSpeech \cite{cheng2020look}, VoxCeleb2 \cite{chung2018voxceleb2}, and TalkingHead-1KH \cite{wang2021one} datasets are given in the Appendix.
Although not perfect, the LRS2 and LRS3 datasets are cleaner than the others by a large margin, and is followed by VoxCeleb2, AVSpeech, and TalkingHead-1KH in the order of decreasing sync quality.
One should therefore remove videos with bad sync quality and restore the sync for tasks that require high AV correlations, such as speech-to-lip synthesis.

\section{Experiment}
\label{sec:experiment}
\subsection{Training details}
\label{ssec:training_details}
\begin{table*}
    \caption{\textbf{Comparison to other methods}. We report results using the test splits. Clip length is the number of consecutive images for a 25 fps video. `Asym' indicates asymmetric image and audio sequence lengths. Number of parameters is in millions (M).}
    \centering
    \begin{tabular}{cccccccccc}
        \toprule
                 &            &        & \multicolumn{6}{c}{Clip length}         & \\
        \cmidrule(lr){4-9}
        Dataset  & Model      & Asym   & 5    & 7    & 9    & 11   & 13   & 15   & $\#$ of params \\
        \midrule
        LRS2     & SyncNet    &        & 75.8 & 82.3 & 87.6 & 91.8 & 94.5 & 96.1 & 13.6M \\
                 & PM         &        & 88.1 & 93.8 & 96.4 & 97.9 & 98.7 & 99.1 & 13.6M \\
                 & AVST       &        & 91.9 & 97.0 & 98.8 & \textbf{99.6} & \textbf{99.8} & \textbf{99.9} & 42.4M \\
                 & VocaLiST   &        & 92.8 & 96.7 & 98.4 & 99.3 & 99.6 & 99.8 & 80.1M \\
                 & ModEFormer &        & 94.5 & 97.1 & 98.5 & 99.3 & 99.7 & 99.8 & 59.0M \\
                 & Ours       & \cmark & \textbf{96.5} & \textbf{98.3} & \textbf{99.2} & \textbf{99.6} & \textbf{99.8} & \textbf{99.9} & 40.6M \\
        \midrule
        LRS3     & AVST       &        & 77.3 & 88.0 & 93.3 & 96.4 & 97.8 & 98.6 & 42.4M \\
                 & ModEFormer &        & 90.9 & 93.1 & 96.0 & 97.7 & 98.7 & 99.2 & 59.0M \\
                 & Ours       & \cmark & \textbf{93.8} & \textbf{96.8} & \textbf{98.3} & \textbf{99.1} & \textbf{99.6} & \textbf{99.8} & 40.6M \\
        \bottomrule
    \end{tabular}
    \label{tab:comparison}
\end{table*}
The LRS2 dataset \cite{afouras2018deep} consists of pretrain, train, validation, and test videos.
We train our model on the pretrain split of the LRS2 dataset for 600 epochs with 50 additional epochs for tuning the batch normalization layers. 
We use batch size of 256, learning rate of $10^{-4}$, and the Adam optimizer \cite{kingma2014adam} with the default beta values.
We use 15 hard negative and 15 easy negative pairs per positive pair, with $w_e=0.1$.

We compare our final accuracy with other methods in Table \ref{tab:comparison}.
We follow the standard practice described in \cite{kadandale2022vocalist} to compute the sync accuracy.
For each video clip, we slide the audio sequence with respect to the images by $(-15,-14,...,0,...,14,15) \times \frac{1}{25} \textrm{sec}$ to obtain $31$ audio sequences with temporal offsets.
We then compute the cosine similarities between the shifted audio sequences and the images, and search for the audio sequence with the minimum similarity.
If this audio sequence has offset within the range $\pm 1$, it is determined to be in sync.
The average number of correct predictions across the test dataset corresponds to the accuracy for clip length $= 5$ in Table~\ref{tab:comparison}.
For larger clip lengths, we compute the average of the cosine similarity for the image sequences falling within the clip length before computing the offset.
We note that for AVST, the evaluation scheme is slightly different in that the model sees all of the frames (instead of taking the average), which is why the increase in accuracy is larger than other methods as clip length is increased.

We also tested our method on the LRS3 dataset~\cite{afouras2018lrs3}, which consists of 118,516 pretrain, 31,982 trainval, and 1,321 test videos.
The model is trained using exactly the same setting as the LRS2 datset.
The results are shown in Table~\ref{tab:comparison}.

\subsection{Ablation study}
\label{ssec:ablation}
\begin{table*}
    \caption{\textbf{Ablation studies}. We report results on the LRS2 validation split. The first line corresponds to the full training scheme. Notations are as follows: EP = total training epochs, DT = drop-and-tune strategy, DB = enable DropBlock, $N_e$ = (average) number of easy negatives per images, $N_h$ = (average) number of hard negatives per images, $N_p$ = (average) number of positive pairs per images, $T_a$ = temporal window of mel spectrogram. Note that the BCE loss corresponds to $N_p=N_h=0.5$.
    }
    \centering
    \begin{tabular}{ccccccccccccc}
        \toprule
        \multicolumn{7}{c}{Training method} & \multicolumn{6}{c}{Clip length} \\
        \cmidrule(lr){1-7} \cmidrule(lr){8-13}
        EP  & DT     & DB     & $N_e$  & $N_h$  & $N_p$ & $T_a$ & 5    & 7    & 9    & 11   & 13   & 15   \\
        \cmidrule(lr){1-13}
        650 & \cmark & \cmark & 15     & 15     & 1     & 32    & \textbf{95.0} & \textbf{97.7} & \textbf{98.8} & \textbf{99.4} & \textbf{99.8} & \textbf{99.9} \\
        350 & \cmark & \cmark & 15     & 15     & 1     & 32    & 94.2 & 97.4 & 98.7 & \textbf{99.4} & 99.7 & \textbf{99.9} \\
        300 & \xmark & \cmark & 15     & 15     & 1     & 32    & 93.5 & 96.8 & 98.4 & 99.2 & 99.6 & 99.8 \\
        300 & \xmark & \xmark & 15     & 15     & 1     & 32    & 93.5 & 96.8 & 98.4 & 99.2 & 99.6 & 99.9 \\
        300 & \xmark & \xmark & 0      & 15     & 1     & 32    & 93.7 & 97.0 & 98.5 & 99.3 & 99.7 & 99.8 \\
        300 & \xmark & \xmark & 0      & 1      & 1     & 32    & 92.2 & 96.1 & 98.0 & 99.0 & 99.5 & 99.8 \\
        300 & \xmark & \xmark & 0      & 0.5    & 0.5   & 32    & 89.7 & 94.6 & 97.1 & 98.4 & 99.2 & 99.6 \\
        300 & \xmark & \xmark & 0      & 0.5    & 0.5   & 16    & 86.2 & 92.6 & 96.3 & 98.2 & 99.1 & 99.5 \\
        \bottomrule
    \end{tabular}
    \label{tab:ablation}
\end{table*}
For ablation studies, we train only for 300 epochs, which is often sufficient to determine the effects of the training strategies.
The results are shown in Table \ref{tab:ablation}.
As can be seen, using asymmetric image and audio sequence lengths, using the BBCE loss with multiple hard negatives, and the drop-and-tune strategy are all important in reducing the error rate.
We note that using small amount of easy negatives can be marginally helpful when the model is trained to saturation, which is why we kept $w_e=0.1$ and $N_e=15$.
However this is not an important component of our training scheme.
In fact, for training for 300 epochs, it seems to slightly hurt performance.

\subsection{Comparison to InfoNCE on LRS2 and LRS3 datasets}
\label{ssec:cf_infonce}
\begin{table*}
    \caption{\textbf{Comparison of BBCE and InfoNCE.} We report results on LRS2 validation set and LRS3 test set.}
    \centering
    \begin{tabular}{cccccccccccccc}
        \toprule
        \multicolumn{7}{c}{Method} & \multicolumn{6}{c}{Clip length} \\
        \cmidrule(lr){1-8} \cmidrule(lr){9-14}
        Loss    & Dataset & DT     & DB     & $N_e$  & $N_h$  & $N_p$ & $T_a$ & 5    & 7    & 9    & 11   & 13   & 15   \\
        \cmidrule(lr){1-14}
        BBCE    & LRS2    & \xmark & \xmark & 0      & 15     & 1     & 32    & 93.7 & 97.0 & 98.5 & 99.3 & 99.7 & 99.8 \\
        InfoNCE & LRS2    & \xmark & \xmark & 0      & 15     & 1     & 32    & 93.4 & 96.8 & 98.4 & 99.2 & 99.6 & 99.8 \\
        BBCE    & LRS3    & \xmark & \xmark & 0      & 15     & 1     & 32    & 92.5 & 96.1 & 97.9 & 98.9 & 99.5 & 99.7 \\
        InfoNCE & LRS3    & \xmark & \xmark & 0      & 15     & 1     & 32    & 92.3 & 95.8 & 97.8 & 98.8 & 99.4 & 99.7 \\
        \bottomrule
    \end{tabular}
    \label{tab:bbce_vs_infonce}
\end{table*}

\begin{table*}
    \caption{\textbf{Comparison of the effects of $N_e$ on InfoNCE and BBCE.} We report results on LRS2 validation set and LRS3 test set.}
    \centering
    \begin{tabular}{cccccccccccccc}
        \toprule
        \multicolumn{2}{c}{Method} & \multicolumn{6}{c}{InfoNCE} & \multicolumn{6}{c}{BBCE} \\
        \cmidrule(lr){1-2} \cmidrule(lr){3-8} \cmidrule(lr){9-14}
        Dataset & $N_e$ & 5    & 7    & 9    & 11   & 13   & 15   & 5    & 7    & 9    & 11   & 13   & 15   \\
        \cmidrule(lr){1-14}
        LRS2    & 0     & 94.1 & 97.3 & 98.6 & 99.3 & 99.7 & 99.8 & 93.8 & 97.1 & 98.6 & 99.3 & 99.7 & 99.9 \\
                & 15    & 94.4 & 97.3 & 98.7 & 99.4 & 99.8 & 99.9 & 93.8 & 97.0 & 98.5 & 99.3 & 99.7 & 99.9 \\
                & 30    & 94.2 & 97.2 & 98.6 & 99.4 & 99.7 & 99.9 & 93.5 & 96.9 & 98.5 & 99.3 & 99.7 & 99.9 \\
                & 60    & 94.2 & 97.3 & 98.7 & 99.4 & 99.7 & 99.9 & 93.4 & 96.9 & 98.5 & 99.3 & 99.6 & 99.8 \\
                & 120   & 94.1 & 97.2 & 98.6 & 99.3 & 99.7 & 99.9 & 93.4 & 96.9 & 98.5 & 99.3 & 99.7 & 99.8 \\
        \cmidrule(lr){1-14}
        LRS3    & 0     & 92.0 & 95.8 & 97.8 & 98.9 & 99.4 & 99.7 & 92.6 & 96.2 & 97.9 & 98.9 & 99.4 & 99.7 \\
                & 15    & 92.6 & 96.2 & 97.9 & 98.9 & 99.4 & 99.7 & 92.4 & 96.0 & 97.8 & 9.88 & 99.4 & 99.7 \\
                & 30    & 92.8 & 96.2 & 98.0 & 99.0 & 99.5 & 99.7 & 92.5 & 96.0 & 97.7 & 98.9 & 99.4 & 99.7 \\
                & 60    & 92.9 & 96.2 & 97.9 & 99.0 & 99.5 & 99.7 & 92.3 & 95.8 & 97.7 & 98.8 & 99.4 & 99.7 \\
                & 120   & 92.8 & 96.3 & 98.0 & 99.0 & 99.5 & 99.7 & 92.2 & 95.8 & 97.7 & 98.8 & 99.4 & 99.7 \\
        \bottomrule
    \end{tabular}
    \label{tab:easy_negatives_combined}
\end{table*}
Similarly to the BBCE loss, we train using the InfoNCE loss on the LRS2 and LRS3 datasets for 300 epochs.
For both LRS2 and LRS3, we find that BBCE performs slightly better than the InfoNCE loss for the 300 epoch setting, see Table~\ref{tab:bbce_vs_infonce}.
Although they perform similarly, the latent space learned by the InfoNCE and BBCE are very different.
For the LRS2 dataset, the InfoNCE setting gives average cosine similarities of approximately 0.47 for positive pairs and 0.14 for negative pairs.
Clearly, there is no natural way to distinguish positive and negative pairs using the cosine similarities.
On the other hand, the InfoNCE loss is directly suited for the model evaluation scheme in Sec.~\ref{ssec:training_details}, since we evaluate whether the model can distinguish a single in-sync pair out of multiple out-of-sync and single in-sync pair.

We observe an interesting behavior when the models are trained further using easy negatives.
In Table~\ref{tab:easy_negatives_combined}, we show the results when the models obtained in Table~\ref{tab:bbce_vs_infonce} are trained further by including various numbers of easy negatives.
For the LRS2 dataset, InfoNCE performs slightly better when using $N_e=15$, but the trend does not continue when using a larger $N_e$.
In contrast, performance on the LRS3 dataset improves significantly when a larger $N_e$ is used. 
In fact, the accuracy is lower when trained to 350 epochs as opposed to 300 epochs for $N_e=0$.
For the  BBCE loss, we trained the models using $w_e=\frac{N_e}{N_e+N_h}$ for analogy.
For both LRS2 and LRS3 datasets, easy negatives were not helpful.
We attribute the difference between InfoNCE and BBCE loss to the fact that InfoNCE loss maximizes a lower bound on the mutual information between audio and visual pairs, whereas BBCE loss maximizes a lower bound  on the mutual information between the AV pairs and sync-labels.
Because LRS3 dataset is collected from TED talks (restricted environment), the mutual information between audio and visual pairs likely contains more spurious correlations when compared to the LRS2 dataset, which is collected from BBC television shows (diverse environment).
Using easy negatives for InfoNCE likely helps combat these spurious correlations, see the Appendix for further discussion.

\section{Conclusion}
In this work, we trained a powerful convolutional sync-net whose output has a probabilistic interpretation.
Because the model is fully convolutional, spatial information is preserved.
Because the loss is based on the BCE loss, the model's output can be interpreted as the probability that the audio and visual segments are in sync.
This property allows the model to be deployed for computing the sync-loss, finding the active speaker, and evaluating AV speech dataset quality.

On the other hand, we have assumed throughout that we have a clean, in-sync dataset.
However, the LRS2 dataset is not perfect, and most of the other AV speech datasets are even dirtier.
Therefore, in order to learn high-quality AV correlations, one must not only filter out various noise such as off-the-screen spearkers, stuttering videos, and dubbed videos, but also bring the audio and visual segments to have as consistent offset as possible. 
Since videos dealt with in such a task are not in sync, one should also consider the robustness of audio encoder to small time translations, which was not considered in this work.
Another problem that we have not discussed in detail is the presence of silent regions during which nobody is speaking.
In principle, any mouth movement is possible in these silent regions, and the model prefers to classify AV pairs from these regions as positives, see the Appendix for further details.
Although it is not immediately clear what the optimal way to treat these silent regions is, we think that an excess of silent regions in the data could be harmful.
We also have not discussed the many-to-many nature of the correspondence between the mouth movements and audio:
our model is trained with the hope that the model will learn to ignore information that cannot be used to infer about the other modality such as identity, but it could be beneficial to consider training schemes that explicitly disentangle these information.
Finally, we have not evaluated the effects of the design choices of the IC-SyncNet on the speech-to-lip generation task.
Because there are numerous variants of the sync-nets and the sync-loss, but a lack of study on the effects of these design choices, a careful evaluation is necessary.
We leave these problems for future research.

\clearpage
{\small
\bibliographystyle{ieee_fullname}
\bibliography{references}
}

\clearpage
\begin{appendices}
\section{Overview}
\label{app:overview}
In Sec.~\ref{app:gradients}, we compare the gradients of the InfoNCE loss and the BBCE loss.
Both are weighted sum of the gradient of cosine similarities, where the weights are given by the probability of mis-classification.
The probability of misclassification indicates that negative pairs are more easily mis-classified, whereas the magnitude of the average cosine similarities of the negative pairs are larger.
We attribute this to the presence of silent regions.
In Sec.~\ref{app.mutual_information}, we discuss the relationship between the BBCE and InfoNCE losses to the lower bounds of mutual information.
We explain how the experiments in the main text are consistent with this interpretation.
In Sec.~\ref{app:active_speaker}, we give the details of the algorithm used to find the active speaker in the main text.
Using the insights gained from this algorithm, we propose in Sec.~\ref{app:lipsync_quality} to use the offset, offscreen ratio, and probability at offset as metrics for lip-sync quality.
We evaluate the quality of various AV speech datasets.

\begin{table*}
    \caption{Notations}
    \centering
    \begin{tabular}{ll}
        \toprule
        $C$ & video clip \\
        $A$ & mel spectrogram \\
        $V$ & image sequence \\
        $f_a$ & audio encoder \\
        $f_v$ & image sequence encoder \\
        $a$ & embedding $f_a(A)$ \\
        $v$ & embedding $f_v(V)$ \\
        $\phi(a,v)$ & cosine similarity $\cos (a,v)$ \\
        $\tau$ & temperature \\
        $\mathbb{E}_X$ & expectation with respect to X \\
        $P(X)$, $P(Y)$, $p(y)$ & probability distributions ($p(y)$: model estimate of $P(Y)$) \\ 
        $H(X)$ & entropy $-\sum_X P(X) \ln P(X)$ \\
        $H(X\vert Y)$ & conditional entropy $-\sum_{X,Y} P(X,Y) \ln P(X\vert Y)$ \\
        $I(X;Y)$ & mutual information $\sum_{X,Y} P(X,Y) \ln \frac{P(X,Y)}{P(X)P(Y)}$ \\
        $D_{KL} (Y \Vert y)$ & KL divergence $\sum_{Y} P(Y) \ln \frac{P(Y)}{p(Y)}$ \\
        $H(Y;y)$ & cross entropy $\sum_Y -P(Y) \ln p(Y)$ \\
        \bottomrule
    \end{tabular}
    \label{tab:notation}
\end{table*}

\section{BBCE as an alternative to InfoNCE}
\label{app:gradients}
Here, we compare the BBCE and the InfoNCE loss in terms of their gradients with respect to the trainable parameters.
To simplify notations, let us define $\phi^+ = \phi(v,a^+)/\tau$ and $\phi^-_i = \phi(v,a^-_i)/\tau$, where $v$ is the image sequence embedding, $a^+$ is the mel spectrogram embedding of a positive pair, $a^-_i$ is the mel spectrogram embedding of the $i$th negative pair, and $\tau$ is the temperature.
We denote by $x$ the trainable parameters involved in computing the similarity metric $\phi$.
The InfoNCE loss is given by
\begin{equation}
    \mathcal{L}_{\textrm{InfoNCE}} = -\ln \frac{\exp(\phi^+)}{\exp(\phi^+)+\sum_{i=1}^{N} \exp(\phi^-_i)},
\end{equation}
where we have omitted the average over mini-batch, and $N$ is the number of negative samples.
Its gradient with respect to $x$ is then given by
\begin{equation}
    \frac{\partial \mathcal{L}_{\textrm{InfoNCE}}}{\partial x} = -p^- \frac{\partial \phi^+}{\partial x} + \sum_{i=1}^{N} p^-_i \frac{\partial \phi^-_i}{\partial x}. \label{eq.infonce_grad}
\end{equation}
Here, $p^- = 1-p^+$,  where $p^+=\frac{\exp(\phi^+)}{\exp(\phi^+)+\sum_{i=1}^N \exp(\phi^-_i)}$ can be interpreted as the probability that $v$ and $a^+$ is a positive pair in the set $\{ a^+, a^-_1,...,a^-_{N}\}$.
Thus, we can interpret $p^-$ as the probability that the model mistakenly classifies the positive pair ($v$, $a^+$) as a negative pair.
Similarly, $p^{-}_i = \frac{\exp(\phi^-_i)}{\exp(\phi^+)+\sum_{i=1}^N \exp(\phi^-_i)}$ can be interpreted as the probability that ($v$, $a^{-}_i$) is mistakenly classified as a positive pair.

The BBCE loss is given by
\begin{equation}
    \mathcal{L}_{\text{BBCE}} = -\ln \frac{\exp(\phi^+)}{1+\exp (\phi^+)} - \frac{1}{N}\sum_{i=1}^N\ln \left(1-\frac{\exp (\phi^-)_i} {1 + \exp (\phi^-_i)}\right).
\end{equation}
For simplicity, we do not distinguish between hard and easy negatives, ignore the one-half factor, and omit average over mini-batch.
Its gradient with respect to $x$ is given by
\begin{equation}
    \frac{\partial \mathcal{L}_{\text{BBCE}}}{\partial x} = -q^- \frac{\partial \phi^+}{\partial x} + \frac{1}{N} \sum^{N}_{i=1} q^-_i \frac{\partial \phi^-_i}{\partial x} . \label{eq.bbce_grad}
\end{equation}
Here, $q^- = 1-q^+$, where $q^+ = \frac{\exp(\phi^+)}{1+\exp (\phi^+)}$ is the probability that ($v$, $a^+$) forms a positive pair.
As before,  $q^-$ has the interpretation as the probability that the model mistakenly classifies ($v$, $a^+$) as a negative pair.
Similarly, $q^-_i = \frac{\exp (\phi^-_i)} {1 + \exp (\phi^-_i)}$ is the probability that ($v$, $a^-_i$) is mistakenly classified as a positive pair.

Note the similarity between \eqref{eq.infonce_grad} and \eqref{eq.bbce_grad}.
For the InfoNCE loss, there is an explicit constraint $p^+ + \sum_{i=1}^{N} p^{-}_i = 1$. 
For the BBCE loss, there is no such explicit constraint, but on average, we find empirically that within a few-percent difference, $q^++q^-_i \approx 1$ for each $i$, so that $q^+ + \frac{1}{N}\sum_{i=1}^{N} q^-_i \approx 1$.
For InfoNCE, $p^- = \sum_{i=1}^{N} p_i^-$, so that the weight given to positive and negative gradients are weighted equally.
For BBCE loss, the weight given to positive and negative losses are weighted equally, so that ideally, we would have a similar relation $q^- \approx \frac{1}{N} \sum_{i=1}^{N} q_i^-$. 
However, the negative and positive pairs are apparently not equally difficult to learn so the ratio $q^- / q^-_i$ can be around $0.7$ \footnote{The precise value can depend on the training method.}. 
This indicates that the negative pairs have larger probability of being mis-classified.
This leads to a larger average weight for the gradients of the negative pairs.

On the other hand, we have commented in the main text that the average cosine similarities of the positive and negative pairs are approximately 0.1 and -0.18, respectively.
At first sight, this seems contradictory: the cosine similarities indicate that negative pairs are correctly classified with a larger margin, and therefore, should be easier to learn.
One source of this behavior is the presence of ``silent regions" during which nobody speaks.
For such regions, any mouth shape is possible, so that when the audio is sampled from these silent regions, it is difficult, if not impossible, to determine the sync (see Fig.~\ref{fig:fig3}) (b) in Appendix~\ref{app:active_speaker} for an example).
If the model classifies (1) non-silent regions with high confidence and high accuracy for negative pairs (2) non-silent regions with lower confidence but high accuracy for positive pairs (3) silent regions as positive pair, the seemingly contradictory behavior can be explained.
Each of these conditions is true for the IC-SyncNet, and although we have not rigorously verified that silent regions are the only culprit, the above observations indicate that silent regions may require special treatment for better performance of sync-nets, especially when trained with the BBCE loss.

To summarize,
\begin{itemize}
    \item The gradients of both the InfoNCE and the BBCE losses are weighted combinations of the gradients of the similarity score.
    \item For the InfoNCE loss, the gradients of the positive pair and the negative pairs are equally weighted. For BBCE, the losses for the positive pair and the negative pairs are equally weighted, but the gradients can be weighted differently depending on the difficulty of learning positive and negative pairs (we find empirically that negative pairs are harder to learn).
    \item Due to the relation $p^+ + \sum_{i=1}^{N} p^{-}_i = 1$ for InfoNCE and the approximate relation (within an order of magnitude) $q^- \approx \frac{1}{N}\sum_{i=1}^{N} q_i^-$, the order of magnitude of the weights given to the gradients of the similarities are the same for InfoNCE and BBCE when the probabilities of classification for InfoNCE and BBCE losses are similar.
    \item Even though negative pairs are classified with a larger margin than positive pairs (on average), the probability of falsely classifying negative pairs as positive pairs is larger than the probability of falsely classifying positive pairs as negative pairs. The silent regions are a source of this inflation of false positives.
\end{itemize}

\section{BCE loss, InfoNCE loss, and mutual information}
\label{app.mutual_information}
Let $A$ and $V$ be the audio and image sequences, let $f_a$ and $f_v$ be the encoders for audio and image sequences, and let $a = f_a(A)$ and $v=f_v(V)$ be the corresponding embeddings.
Let $Y$ be the (ground-truth) label of whether $A$ and $V$ are in sync.
The predicted probability that $A$ and $V$ are in sync is modeled by $p(Y \vert a,v)=\sigma(\phi(a,v)/\tau)$.
The cross entropy loss is 
\begin{align}
    H(Y;y\vert a,v) &= -\sum_{Y,a,v} P(Y,a,v) \ln p(Y\vert a,v) \\
    &= H(Y\vert a,v) + D_{KL}(Y \Vert y\vert a,v),
\end{align}
where 
\begin{align}
H(Y\vert a,v) &= -\sum_{Y,a,v} P(Y,a,v) \ln P(Y\vert a,v), \\
D_{KL}(Y \Vert y \vert a,v) &= \sum_{Y,a,v}P(Y,a,v)\ln \frac{p(Y|a,v)}{P(Y|a,v)}.
\end{align}
The mutual information between $Y$ and $a,v$ is
\begin{equation}
    I(a,v;Y) = H(Y) - H(Y\vert a,v).
\end{equation}
Thus, we have
\begin{align}
    H(Y;y\vert a,v) &= H(Y) - I(a,v; Y) + D_{KL}(Y \Vert y\vert a,v) \\ 
    &\ge H(Y) - I(a,v ; Y), \label{eq:bbce_mutual_info}
\end{align}
where the second line uses the fact that the $D_{KL} \ge 0$.
From the first line, we see that because $H(Y)$ is fixed (entropy of labels), minimizing the cross entropy loss maximizes the mutual information between $a,v$ and the in-sync labels, while also aligning the distributions $P(y\vert a,v)$ with $P(Y\vert a,v)$.
Another way of interpreting the result is that minimizing the cross entropy loss amounts to maximizing a lower bound on the mutual information $I(a,v;Y)$. 
Because $I(a,v;Y) \le I(A,V;Y)$, which can be shown by using the data processing inequality on the Markov chain $Y \rightarrow (A,V) \rightarrow (a,v)$, we are in effect, maximizing the lower bound on $I(A,V;Y)$.
Since the loss function used in our paper is nothing but this cross entropy loss with balanced positive and negative samples, we are in effect maximizing a lower bound on the mutual information between $A,V$ and the labels $Y$.
We note that similar discussion was also given in the context of metric learning, see \cite{boudiaf2020unifying}. 

It is well known that the InfoNCE loss is also related to the mutual information.
By following the arguments in \cite{oord2018representation}, we see that minimizing the InfoNCE can be interpreted as maximizing a lower bound on the mutual information between $a$ and $v$ when a large number of negatives are used.
For completeness, we repeat the argument in \cite{oord2018representation} using the notation relevant to the problem considered in this work.
First, note that the InfoNCE takes the form
\begin{equation}
    \mathcal{L}_{\textrm{InfoNCE}} = - \mathbb{E}_{v,a_1,...,a_N} \left[ \frac{f(a_i, v)}{f(a_i,v) + \sum_{j\neq i}^{N} f(a_j,v)}\right].
\end{equation}
Without loss of generality, we can choose  $i=1$ to be the positive sample (i.e. in-sync with $v$) and $a_2,...,a_N$ to be the negative samples from the distribution $p(a)$.
The loss can be interpreted as the cross-entropy loss of classifying $a_1$ as the positive sample, where $\frac{f}{\sum{f}}$ is the probability that $a_1$ is a positive sample out of the distribution followed by $p(a)$.
In this setting, the probability that optimizes the loss function is given by
\begin{equation}
    P = \frac{p(a_1\vert v) \prod_{i\neq 1} p(a_i)}{\sum_{i=1}^{N}p(x_i\vert v)\prod_{j\neq i} p(a_j)},
\end{equation}
so that the optimal values of $f$ that minimize the loss is given by $f(a,v) \propto \frac{p(a\vert v)}{p(a)}$.
Note that this is implemented using the parametrization $f(a,v) = \exp \left( \phi(a,v) / \tau \right)$.
Assuming that $N$ is large, we can rewrite the InfoNCE loss at the optimal $f$ as follows:
\begin{align}
    \mathcal{L}^{\textrm{opt}}_{\textrm{InfoNCE}} &= \mathbb{E}_{v,a_1,...,a_N} \ln \left[ 1 + \frac{p(a_1)}{p(a_1\vert v)} \sum_{i=2}^{N} p(a_i\vert v) / p(a_i)\right] \\
    &\approx  \mathbb{E}_{v,a_1} \ln \left[ 1 + (N-1)\frac{p(a_1)}{p(a_1\vert v)}\right]  \\
    &\ge \mathbb{E}_{v,a_1} \ln \left[N\frac{p(a_1)}{p(a_1\vert v)}\right] \\ 
    &= -I(a_1;v) + \ln N. \label{eq:infonce_mutual_info}
\end{align}
A drawback of this derivation is that it uses the approximation that $N$ is large.
A more sophisticated derivation can be given without this assumption, see \cite{poole2019variational}.

Now, let us consider a dataset consisting of a single (large) video clip $C$ of a person giving speech.
Here, we randomly draw multiple $v$'s (mini-batch), and for each $v$, we draw one positive $a^+$ and multiple random $a^-_i$'s.
For this case, we can interpret $p(a)$ and $p(v)$ as the probability distributions of audio and image segments for this single video.
Then, minimizing the InfoNCE loss $\mathcal{L}_{\textrm{InfoNCE}}(C)$ amounts to maximizing the lower bound on the mutual information between $a$ and $v$ of this video clip.
Next, let us consider where we have multiple videos of various persons giving speech.
Now, the randomly chosen $v$'s are from different videos and $a^+$'s are from the same video as $v$'s.
Let us consider the case where $a^-_i$ are from the same video as $v$.
Then, the InfoNCE loss over the whole dataset is $\mathcal{L}_{\textrm{InfoNCE}} = \mathbb{E}_{C} \mathcal{L}_{\textrm{InfoNCE}}(C)$.
Thus, we are maximizing the lower bound on the average of mutual information $I(a;v)$ for $(A,V) \in C$: $\mathcal{L}_{\textrm{InfoNCE}} \ge -\mathbb{E}_{C}I(a;v)$, where $(A,V) \in C$.
Finally, let us consider the case where $a^-_i$ are chosen randomly from different videos in the InfoNCE.
Then, we are maximizing the mutual information between $a$ and $v$, where $A$ and $V$ are from random frame in random videos. \footnote{Note that we ignore the stochasticity of loss minimization in our arguments.}

Now, let us consider the InfoNCE for the second case discussed above.
The InfoNCE loss has the effect of training the model to identify the positive audio sample out of multiple hard negative audio samples, but it also maximizes the lower bound on the mutual information between the audio and images in the same video.
This can nudge the model to learn various correlations that are not directly related to AV sync.
If the negatives are chosen at random across all of the videos as in the third (final) case discussed above, this has the effect of training the model to maximize the mutual information between the image and the audio from other videos.
Since different videos are from different identity, this can induce the model encode identity information in $a$ and $v$.
The optimal sampling scheme seems to lie somewhere between these two extremes, where one uses both hard and easy negatives.
Note, however, that for such sampling schemes, there is no simple relation to the mutual information \cite{tschannen2019mutual}.

In ModEFormer, the authors propose one way to do this via the two-stage training strategy, where during the first stage, only hard negatives are used, and during the second stage, both hard and easy negatives are used.
We suspect that the spurious correlations learned during the first stage that help identify correct pairs are forgotten during the second stage because these correlations are no longer helpful in identifying the correct pair in the presence of data from other videos.
In the main text, we tested this behavior on the LRS2 and LRS3 datasets by first training the model with 15 hard negatives with the InfoNCE loss for 300 epochs, and then fine tuning for additional 50 epochs with various numbers of easy negatives.
As was shown in Table 4 in the main text, $N_e=N_h$ performed the best for LRS2 dataset, but using a larger number of easy negatives was helpful for LRS3 dataset.
Furthermore, training for 350 epochs with InfoNCE without using any easy negatives was helpful for the LRS2 dataset, but lead to overfitting for the LRS3 dataset.
We believe that this difference results from the fact that LRS3 dataset can contain more spurious correlations than the LRS2 dataset because the videos in the LRS3 dataset are collected from a restricted environment.
Such correlations are difficult to identify: the mic, which is often visible, could be a culprit, or there could be some correlation between the head pose and the audio.
However, it is clear that the possibility of there being spurious correlations is larger for such a biased dataset.

To study whether the BBCE loss shows a similar behavior, we trained our model with the BBCE loss for 300 epochs using only the hard negatives, and further tuned the model using hard and easy negatives for additional 50 epochs.
For analogy to the InfoNCE loss, we used a proportional weighting scheme for the easy negatives, meaning that $w_e = \frac{N_e}{N_e+N_h}$.
The results in Table 4 in the main text indicate that easy negatives are not helpful for the BBCE loss for both LRS2 and LRS3 datasets with this proportional weighting scheme.
We believe that this is because the BBCE loss trains the model to learn information directly related to AV sync (see Eq.~\eqref{eq:bbce_mutual_info}) so that there is less need for regularization.

\section{Active speaker detection}
\label{app:active_speaker}
It is possible to use the IC-SyncNet for active speaker detection.
More precisely, it can be used to determine whether there is a meaningful correlation between the mouth images and the audio, which is near frame-wise accurate.
A naive approach to this problem would be
\begin{enumerate}
    \item Compute the offset
    \item Synchronize the audio and images using the offset
    \item Use the threshold $p=0.5$ to determine whether the audio and images are in-sync for each frame
\end{enumerate}
However, this gives noisy results because the model's output is not robust to small translation in the time direction. Thus, a slightly more sophisticated approach is necessary.

To proceed, let $S[i,j]=\hat{a}_i \cdot \hat{v}_j$ be the cosine similarity matrix.
Here, $j=0,1,...,N_{v}-1$ is the index of image sequence in the video and $i=j-w,j-w+1,...,0,...,j+w-1,j+w$ is the index of the audio window of size $2w+1$ around the image at time $j$.
Let $i=m$ be the location where the average of $S[i,j]$ along the second axis is maximum. Then, $w-m$ is the offset between the audio and images.
Before moving on, let us note that the time interval between two consecutive indices in the audio window is approximately $1/25$ seconds.
We say approximately because the time step in the mel spectrogram is not commensurate with that of images (the ratio of the interval between images to mel spectrograms is $3.2$), so one must make approximations.
It seems to be common practice (see for instance the VocaLiST \href{https://github.com/vskadandale/vocalist/blob/d2d7d4fe2df03a9ad7b36d93cdf22dee1a6f0217/test_lrs2.py#L160}{code}) to compute this by $i = \textrm{floor}(3.2 \times j)$.
This is suboptimal (one should round to the nearest integer) but we follow the conventional practice throughout this paper.

The active speaker detection is carried out as follows.
\begin{enumerate}
    \item Pass $S[i,j]$ through a 1D Gaussian filter along the first dimension. We use filter size = $3$.
    \item Compute the probability matrix $P[i,j] = \sigma \left( S_f[i,j] / \tau \right)$, where $S_f[i,j]$ is the smoothed cosine similarity.
    \item Compute the maximum probability $P_m$ in a small window around $m$: $P_m[j] = \max P[m-1:m+1,j]$
    \item Obtain candidate on-screen regions through thresholding: $P_t = (P_m > 0.4)$
    \item Get connected components of $P_t$. These are denoted by $(C_k,B_k)$. Here, $C_k$ is an interval such that $P_t[C_k]=B_k$, and $k$ indexes these chunks.
    \item Threshold each of the connected components: if $\textrm{mean} (P_m[C_k]) < 0.66$, we set $B_k=\textrm{False}$. Merge neighboring $C_k$'s which have the same $B_k$'s.
    \item (optional) Remove islands and archipelagos. 
\end{enumerate}
By islands, we refer to small chunk of $C_k$ surrounded by larger chunks.
This can happen for various reasons: it could be just noise, the person may have stopped speaking for a small interval of time, etc.
Therefore, we merge islands into the neighbors (i.e. set $B_k$ equal to $B_{k+1} = B_{k-1}$).
By archipelagos, we refer to the presence of consecutive small chunks ($C_k$).
The presence of archipelagos usually indicate that the region is noisy (model's output is uncertain).
To decrease false positives, we prefer to remove these regions (i.e. set $B_k$'s for archipelagos to False). 
In our implementation, we consider $C_k$ with size less than 4 to be small (the heuristic is rooted in the assumption that $\pm 1$ is a valid noise in offset, which has size 3).
The other thresholds are roughly chosen based on several examples we have examined, and the optimal values will depend on the model, train data, test data, and application.

\begin{figure*}
	\centering
	\includegraphics[width=12cm]{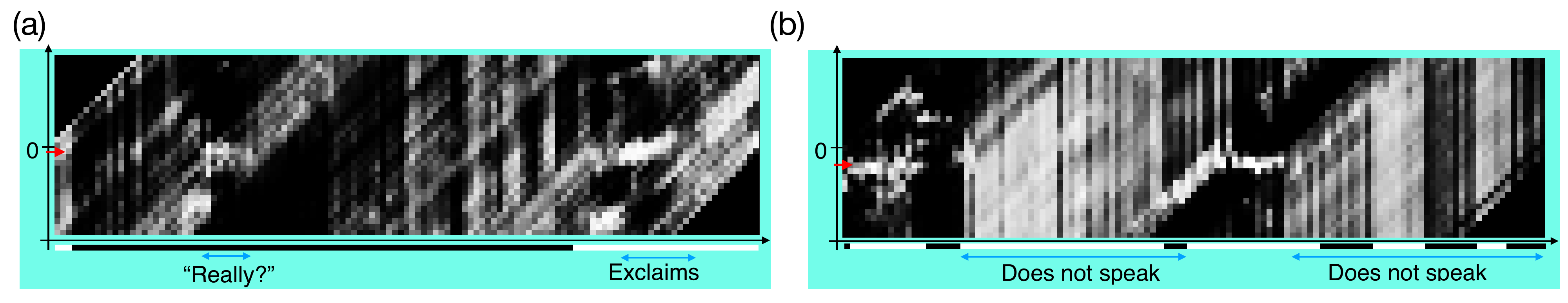}
	\caption{Failure cases. Offset for (a) is -1 and offset for (b) is -3. In (b), the person does not speak for the duration of the parallelogram regions.}
	\label{fig:fig3}
\end{figure*}

We show some failure cases in Fig.~\ref{fig:fig3}. 
In (a), which is from the Voxceleb2 dataset \cite{chung2018voxceleb2}, the segment where the person captured (Allison Scagliotti) says ``really?" is filtered out during step 6 ($\textrm{mean} (P_m[C_k]) < 0.66$).
This can be remedied by using a less strict requirement if desired (we do not like false positives, and prefer to use the stricter requirement).
In (b), which is from the AVSpeech dataset \cite{cheng2020look}, the region where the lecturer does not say anything is determined to be in sync.
This happens because any mouth shape can be possible when one is not speaking, and is a normal (and common) behavior of sync-nets.
Such regions occurs in parallelograms, and can be detected if desired by looking for parallelograms in the sync probability or by using voice activity detection algorithms.

\section{Lip-sync quanlity of AV speech datasets}
\label{app:lipsync_quality}
\begin{figure*}
	\centering
	\includegraphics[width=12cm]{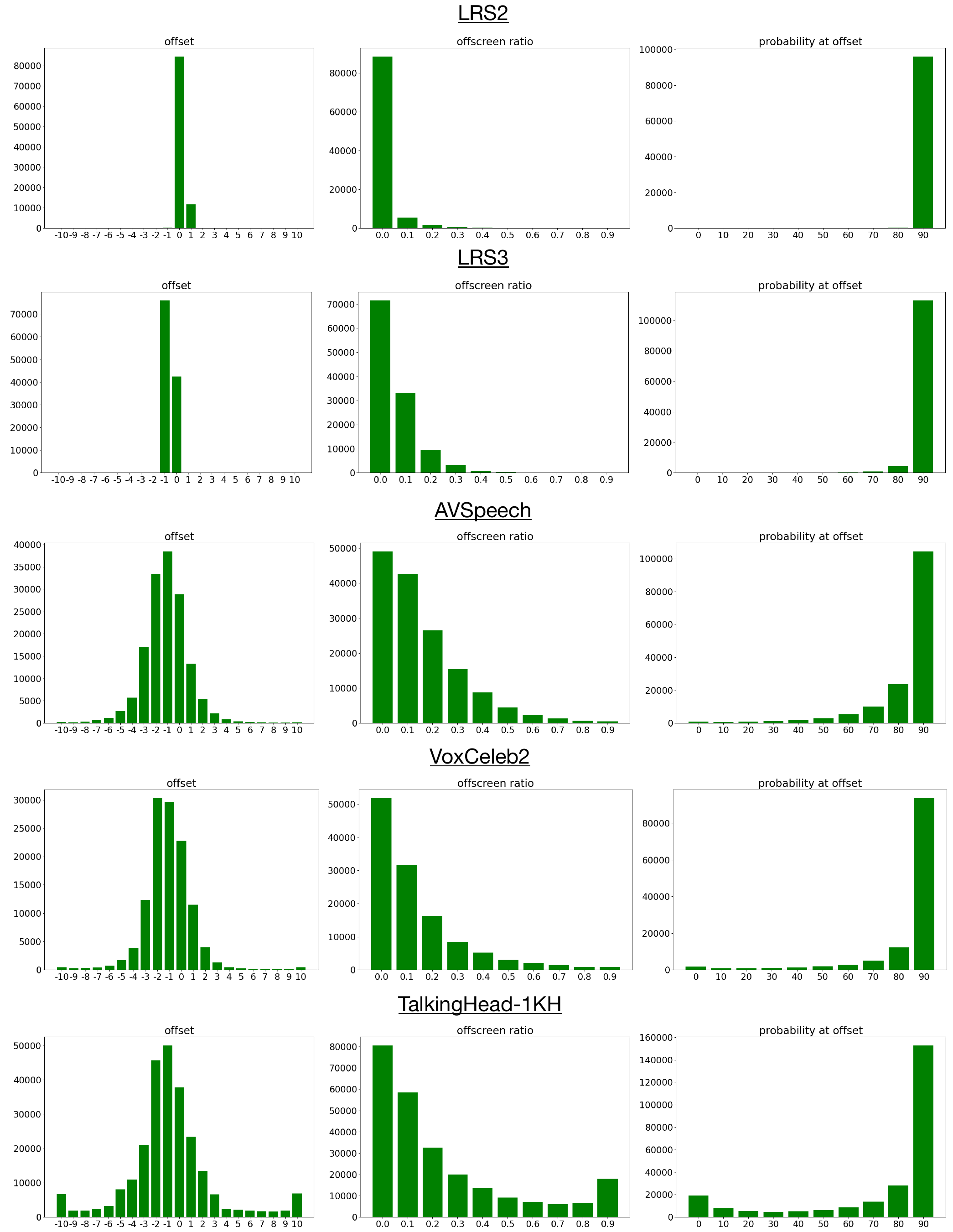}
	\caption{Lip-sync quality of various AV speech datasets.}
	\label{fig:fig4}
\end{figure*}
We show statistics on the sync quality of some of the well known AV speech datasets in Fig.~\ref{fig:fig4}.
Let us first explain the metrics in the histograms.
The offset is computed by finding the best offset value for the whole video.
The offscreen ratio is found by (1) finding the offset for the whole video (2) smoothing the cosine similarity along the $x$ direction as in Appendix~\ref{app:active_speaker} (3) finding the cosine similarity that gives the best in-sync probability in the $\pm 1$ window about the offset for each frame (4) using the threshold of $0.66$ for the probability that the images and audio are in sync (5) and finding the ratio of the frames for which the image and audio sequences are determined not to be in sync (i.e. offscreen).
The probability at offset is the probability of the average of the cosine similarity found in step 3).

As can be seen, there is a large gap in quality  between the LRS datasets (i.e. LRS2 and LRS3) and the others, even if we consider the fact that the model was trained on the LRS2 dataset \footnote{Note that we pre-process the AVSpeech, VoxCeleb2, and TalkingHead-1KH by splitting them into maximum of 5-second chunks, so the number of data can be different.}.
By examining the offscreen ratio, we see that even the LRS2 dataset (pretrain split) contains videos with off-the-screen speakers: 6208622187442383782/00021.mp4 and 6176248871448853539/00036.mp4 are just few of the examples found by looking for videos with offscreen ratio larger than 0.2.
The probability at offset is a slightly different metric, and is correlated with the overall sync quality.
Stuttering videos form one example class of videos that are in sync on average, but has low probability at offset.
Combining the offscreen ratio with probability at offset, it is possible to filter out low quality data.
The offset values for the filtered data could be large, but this is not a problem because the videos can be brought back into sync by using the offset.
The resulting filtered dataset contains higher quality AV correlations than the raw data, and we recommend carrying out such a process to obtain best results, especially on sensitive AV tasks.
Probability at offset = $0.9$ and offscreen ratio=$0.2$ are reasonable values for filtering out bad data.

Another characteristic of the AV speech datasets that can affect the quality of AV correlations is the presence of silent regions.
As discussed in Appendix~\ref{app:gradients}, any realistic mouth shape is possible during these silent regions, so that it is near impossible to determine the sync.
This is the reason that the model prefers to classify AV pairs with audio from these silent regions as positives, as can be seen from Fig.~\ref{fig:fig3} (b).
The optimal way to treat these silent regions is not clear at the moment (e.g. should we sample negatives less from silent regions?).
However, it is unlikely that data for which silent regions form a significant portion is meaningful for AV tasks, and recommend filtering these out as well.


\end{appendices}

\end{document}